# Deep Learning-Based Age Estimation and Gender Classification for Targeted Advertisement

Muhammad Imran Zaman[1], Nisar Ahmed[2]
[1]Department of Computer Science, COMSATS University Islamabad – Lahore Campus, Lahore, Pakistan.
[2]Department of Computer Science (New Campus), University of Engineering and Technology Lahore, Pakistan.
[1]imranzaman.ml@gmail.com, [2]nisarahmedrana@yahoo.com

*Abstract*— This paper presents a novel deep learning-based approach for simultaneous age and gender classification from facial images, designed to enhance the effectiveness of targeted advertising campaigns. We propose a custom Convolutional Neural Network (CNN) architecture, optimized for both tasks, which leverages the inherent correlation between age and gender information present in facial features. Unlike existing methods that often treat these tasks independently, our model learns shared representations, leading to improved performance. The network is trained on a large, diverse dataset of facial images, carefully pre-processed to ensure robustness against variations in lighting, pose, and image quality. Our experimental results demonstrate a significant improvement in gender classification accuracy, achieving 95%, and a competitive mean absolute error of 5.77 years for age estimation. Critically, we analyze the performance across different age groups, identifying specific challenges in accurately estimating the age of younger individuals. This analysis reveals the need for targeted data augmentation and model refinement to address these biases. Furthermore, we explore the impact of different CNN architectures and hyperparameter settings on the overall performance, providing valuable insights for future research.

*Keywords—deep learning, age estimation, gender classification, face image analysis, targeted advertisement.*

## I. INTRODUCTION

The human face serves as a powerful medium for conveying various characteristics, including expression, age, gender, ethnicity, and identity. Leveraging advancements in machine learning and computer vision, researchers have been able to harness facial images to identify and analyze these human traits, giving rise to significant interest in both academic and industrial circles [1-3]. Two key areas of research have emerged: face image synthesis [4, 5] and face image analysis [6, 7].

Face image synthesis involves the development of techniques to generate realistic facial images based on specific attributes such as age, gender, expression, identity, ethnicity, and pose. On the other hand, face image analysis focuses on interpreting facial images to extract information about these attributes. Among the myriad applications of face image analysis; gender identification and age estimation stand out as particularly intriguing methods with wide-ranging implications, including targeted advertising and human-computer interaction.

It is remarkable how humans possess the innate ability to accurately discern gender and estimate age from facial images, even from a young age [8]. However, replicating this capability in machines presents a significant challenge. Age estimation, in particular, is a complex task influenced by various factors such as facial appearance, expression, makeup, and occlusion [9]. Despite these challenges, advancements in machine learning algorithms offer promising avenues for achieving accurate age estimation from facial images.

Aging, a natural and irreversible process, manifests in distinct patterns and variations in facial appearance over time [10]. From changes in facial shape during prepubescence to alterations in skin texture and the emergence of wrinkles in adulthood, the aging process introduces unique challenges for age estimation algorithms [11, 12]. Understanding these aging patterns is essential for developing robust machine learning models capable of accurately estimating age from facial images.

Furthermore, gender identification from facial images has garnered significant attention in both psychology and computer vision [13-16]. While humans excel at distinguishing between male and female faces with remarkable accuracy, achieving similar performance in automated systems remains a challenge. Nonetheless, advancements in object detection, preprocessing, feature extraction, and classification techniques have paved the way for improved gender identification algorithms [17].

The applications of age and gender identification from facial images span diverse domains [18-20], including biometrics, surveillance [21-25], electronic customer relationship management, content access control, and targeted advertising [10, 26]. These technologies offer opportunities for enhancing security, personalizing user experiences, and optimizing business processes.

Although age and gender identification from facial images has advanced, it is still very difficult to achieve human-level accuracy. It will require interdisciplinary cooperation between researchers in computer vision, machine learning, psychology, and related fields to address these challenges. This study explores the intricate nature of age and gender recognition from facial images, looking into the methods, potential uses, and underlying theories. Through improving our understanding of these mechanisms, we hope to encourage the creation of more accurate and robust algorithms with concrete applications in the real world.

## II. RELATED WORK

### A. Introduction to Age and Gender Identification

Age and gender identification from facial images represent critical tasks in computer vision, with applications across various domains necessitating accurate predictions. These tasks are essential for decision-making processes in fields such as security systems, biometric authentication, medical imaging, demographic research, content-based searches, and surveillance systems.

*B. Gender Identification*

Several studies have explored gender prediction from facial images using conventional machine learning methods [27]. These studies typically involve stages such as feature extraction and classification. For instance, one study utilized local binary pattern (LBP) and histogram of gradient (HOG) methods for feature extraction and achieved state-of-the-art results using SVM and KNN classifiers [19, 20, 27]. Another study proposed a framework based on Gabor features [19] and ensemble classifiers [14, 16, 28], achieving high accuracy on multiple datasets [29]. Additionally, researchers have explored the use of convolutional neural networks (CNNs) for gender prediction, developing customized architectures with improved efficiency and accuracy [24].

*C. Age Estimation*

Age estimation from facial images has been addressed using various approaches, including deep learning-based models and multi-stage feature constraint learning frameworks. Some studies have focused on enhancing the efficiency and accuracy of age estimation through lightweight CNN architectures [30]. Others have proposed ensemble classifiers and label distribution learning techniques to handle the complexity of age variations across different demographic groups [31, 32]. Moreover, researchers have explored the integration of gender classification modules to improve the overall performance of age estimation systems [33].

*D. Integration of Age and Gender Identification*

While many studies have focused solely on either gender prediction or age estimation, a limited number have addressed both tasks simultaneously. Integrating age and gender identification presents challenges due to the distinct nature of classification (gender) and regression (age estimation) tasks. However, some approaches have attempted to tackle both tasks within a unified framework, leveraging deep learning models and multi-task learning strategies [34, 35]. Additionally, the incorporation of multimodal data, such as facial and ear images, has shown promise in improving the accuracy of age and gender classification systems [36].

*E. Research Gap*

Despite notable progress in age and gender identification from facial images, there remains a notable gap in the existing literature. Specifically, there is a dearth of comprehensive studies that simultaneously tackle age, gender, ethnicity, and other attributes prediction from facial images. Moreover, the exploration of age variations across different demographic groups, particularly in early childhood and late adulthood scenarios, is not extensively covered.

Furthermore, the integration of gender classification modules into age estimation frameworks has not been thoroughly investigated to enhance system performance. Notably, existing studies on simultaneous age and gender classification are not specifically formulated to address the unique requirements of targeted advertisement.

In summary, while the literature on age and gender identification from facial images demonstrates significant progress and potential, the existing research lacks comprehensive coverage and optimization for targeted advertisement applications. Bridging these gaps through interdisciplinary collaboration and innovative methodologies is essential for advancing the field and unlocking its full potential in various domains, including targeted advertisement.

III. METHODOLOGY

The problem of age and gender identification is pivotal, with applications spanning various domains. Despite numerous attempts by researchers, as outlined in the related work, achieving human-level accuracy in gender identification and age estimation remains elusive. Therefore, there is a pressing need for further research in this area. This chapter outlines the methodology employed in designing and implementing a face image-based age and gender identification system. To ensure a systematic approach and optimal results, a structured methodology for machine learning project development is adhered to. Given the diverse requirements of different machine learning algorithms and the need for meticulous data processing and model evaluation, the adoption of systematic methodologies is imperative. In this study, we have relied on CRISP-DM [37] to serves as a structured workflow for age and gender identification, encompassing six stages from conception to deployment.

*A. Problem Understanding*

Defining the problem is paramount in any machine learning endeavor, laying the foundation for an effective solution. This study aims to conduct age estimation and gender classification based on facial images. Following preprocessing, feature extraction, and model training, a machine learning model will be trained on a face image dataset. Model validation will adhere to predefined evaluation criteria, leading to the deployment of the final model for practical use.

It's crucial to distinguish between age estimation and gender classification, as they represent distinct yet interconnected challenges. Age estimation involves regression, minimizing the disparity between actual and predicted ages using a designated loss function, such as mean squared error. Conversely, gender identification is a classification task, categorizing face images into binary genders. While non-binary gender identification is feasible with sufficient training data, this study focuses on binary classification due to dataset limitations. As both tasks stem from the same dataset, preprocessing steps remain consistent for both solutions. Ultimately, the objective is to accurately predict the age and gender of human subjects.

*B. Dataset Understanding*

The dataset understanding section for this study hinges on the specific requirements of age and gender identification tasks. The UTK Face dataset is chosen due to its comprehensive coverage, containing 20,000 annotated face images. These images span a wide age range from birth to 116 years and encompass various facial expressions, poses, illuminations, occlusions, and resolutions. The dataset amalgamates images from three primary sources: Morph, a longitudinal age dataset comprising 55,134 images of 13,617 individuals aged 16 to 77 years; CACD, a cross-age celebrity dataset with 163,446 images of 2,000 celebrities aged from 2004 to 2013; and additional images collected via web crawling, specifically targeting children, newborns, and older adults.

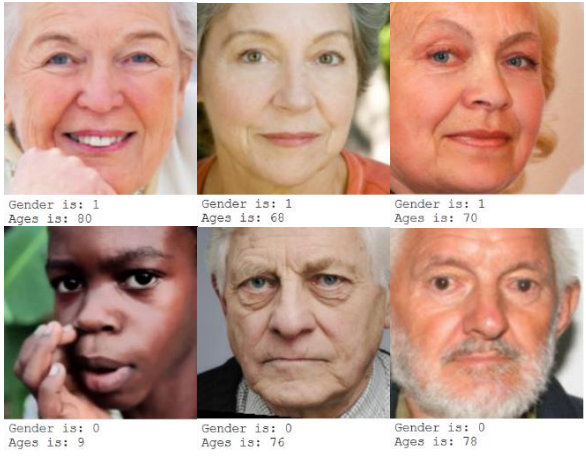

FIGURE 1: SIX SAMPLE IMAGES FROM THE DATASET ALONG WITH THE ASSIGNED AGE AND GENDER LABELS.

## C. Data Preparation

Data preparation is crucial, especially considering the chosen model's requirements. Since CNNs are widely used for age estimation and gender classification, we adopt a CNN-based approach due to its effectiveness in image processing. Despite CNNs' inherent ability to handle image data with minimal preprocessing, adjustments are needed for the UTK Face dataset. A substantial number of images are concentrated in the one to four-year age group (as depicted in Figure 2), potentially biasing model predictions. To mitigate this bias, we randomly sample 20% of the images from this age group, ensuring balanced data distribution.

In gender classification, achieving a balanced number of samples in each class is crucial to avoid issues arising from class imbalance. Figure 3 displays the number of face images in each class, where class label "1" denotes female and "0" denotes male. Notably, a third class labeled "3" is present, likely indicating non-binary gender or an erroneous entry, and images labeled with class "3" are removed from the dataset.

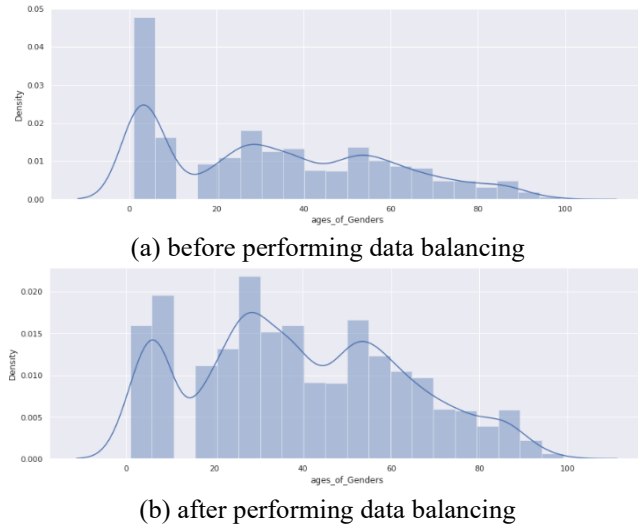

(a) before performing data balancing

(b) after performing data balancing

FIGURE 2 DISTRIBUTION OF AGE BEFORE AND AFTER DATA BALANCING

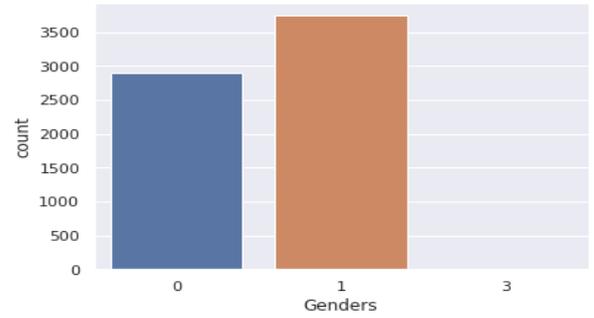

FIGURE 3: NUMBER OF FACE IMAGES IN EACH CLASS.

Given that CNNs require fixed input resolution, all input images are resized to 200×200 pixels. Additionally, feature normalization is applied to enhance training convergence and regularization performance. With 8-bit depth images, pixel values ranging from 0 to 255, feature normalization is conducted using Eq. 1, where $S_i$ represents the original pixel value, and $S_i'$ represents the scaled pixel value.

$$S_i' = \frac{S_i - \min(S_i)}{\max(S_i) - \min(S_i)} \quad (1)$$

## D. Model Building

Model building is a crucial step that integrates problem understanding, data preparation, and data understanding. Given the utilization of CNNs in this process, the data must be formatted appropriately. One common approach for model training and validation is hold-out validation, where the dataset is randomly divided into training and validation sets to prevent bias. In this experiment, with a dataset of 6,647 samples, a 70-30 split is chosen due to the limited size of the dataset.

The model building process involves various steps, including determining the training strategy (either training from scratch or transfer learning), designing and implementing an image augmenter, defining the CNN architecture, conducting hyperparameter search, and finally, model training and validation.

**Training Strategy**

For CNN training, two main strategies are typically employed: training from scratch and transfer learning. Transfer learning involves using a pre-trained CNN model trained on a larger dataset. However, in this study, pre-trained models from ImageNet did not align well with the problem domain of age and gender prediction. Therefore, training the CNN model from scratch was deemed more suitable, given the small size of the UTK Face dataset.

**Data Augmentation**

Data augmentation, a technique to increase the effective size of the dataset by introducing variations, was explored. However, due to the alignment and cropping of images in the UTK Face dataset based on facial landmarks, data augmentation did not yield significant improvements. Introducing perturbations resulted in the loss of facial landmarks, consequently reducing prediction performance.

**CNN Architecture**

The architecture of CNN serves as the backbone of the modeling process, providing a hierarchical structure to learn patterns from images. Unlike fully connected neural networks, CNNs are designed to handle images efficiently by leveraging their inherent redundancies. Regularization techniques such as skip connections, dropout, and weight decay are employed to prevent overfitting. The basic CNN architecture consists of input, convolutional, pooling, and fully connected layers, culminating in an output layer.

In the age estimation architecture, the output layer comprises a single neuron with a ReLU activation function for regression. This neuron maps the ReLU output to the corresponding age value. Conversely, in gender identification architecture, the last fully connected layer contains one neuron followed by a softmax layer for probability computation and classification. Despite slight variations, both architectures follow a similar structure, with the last layer tailored to the specific task of age estimation or gender classification.

The input layer's size is determined by the dimensions of the input feature vector, which, in this case, is an image of 200x200x3 pixels. This architecture ensures efficient learning of patterns from images and facilitates accurate prediction of age or gender. Table 3 presents the architecture used for model building, noting that while it represents the age estimation model, the gender classification model follows a similar structure with adjustments to the output layer.

**Model Training**

Model training is the process in which the designed CNN architectures is trained by supplying the pre-processed training data. The model training for both the problems is performed using the training parameters of Table 4. It is to be noted that hyperparameter search is performed using grid search to identify the optimal configuration of model parameters.

TABLE 1: HYPERPARAMETERS USED IN MODEL TRAINING

| Sr. No. | Parameter | Value |
|---|---|---|
| 1 | Optimizer | Adam |
| 2 | Initial Learning Rate | 0.01 |
| 3 | Momentum | 0.9000 |
| 4 | Max Epochs | 150 |
| 5 | Batch Size | 32 |
| 6 | Shuffle | Every epoch |
| 7 | Regression Loss | MSE |
| 8 | Regression Metric | RMSE |
| 9 | Classification Loss | Cross-entropy |
| 10 | Classification Metric | Accuracy |

Figure 4 shows the training progress of the regression CNN for age estimation from face images. It employs mean square error as the loss function and root mean squared error as the validation metric. Meanwhile, Figure 5 depicts the training progress of the classification CNN for gender prediction from face images. The model utilizes cross-entropy as the loss function and measures validation performance using classification accuracy.

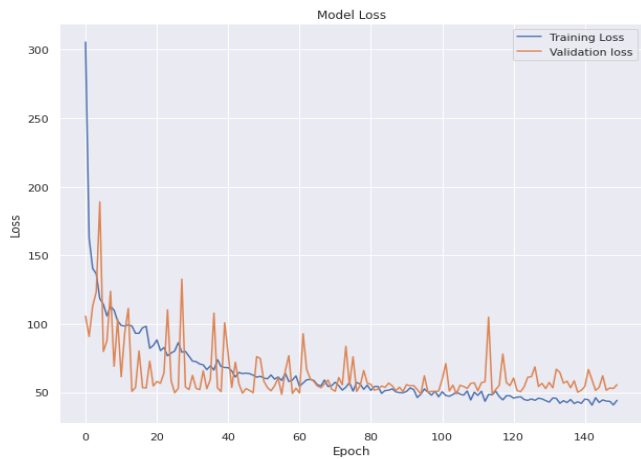

FIGURE 4 TRAINING PROGRESSION OF REGRESSION CNN (AGE ESTIMATION)

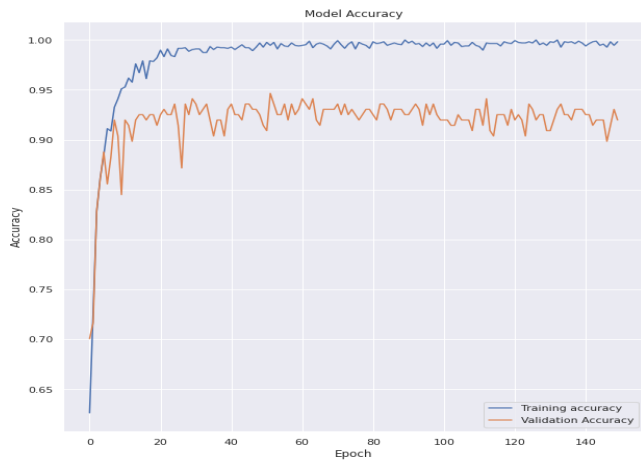

FIGURE 5 TRAINING PROGRESSION OF CLASSIFICATION CNN (GENDER IDENTIFICATION)

IV. EXPERIMENTAL RESULTS

The experimental results section presents the outcomes obtained from the application of the proposed methodologies to address the age estimation and gender prediction tasks using facial images. These results provide insights into the effectiveness and performance of the developed models in accurately estimating ages and predicting genders. Evaluation metrics such as mean squared error and classification accuracy are employed to assess the models' performance, providing a comprehensive analysis of their effectiveness in real-world applications.

*A. Model Evaluation*

Model evaluation quantifies the effectiveness of different model architectures and design decisions. It involves defining evaluation criteria and metrics to measure and compare performance. Two common validation strategies are holdout validation and cross-validation. This study has used holdout validation due to smaller number of samples in the dataset and therefore 30% images are put aside for model testing whereas the rest of 70% images are used for model training and validation.

**Evaluation Metrics**

Evaluation metrics vary depending on the problem type and dataset characteristics. For age estimation, regression metrics such as Root Mean Squared Error (RMSE) and Mean Absolute Error (MAE) are used to validate the performance of age estimation. The formulas for both metrics are provided in Eq. 2 and 3.

$$RMSE = \sqrt{\frac{1}{n}\sum_{i=1}^{n}(y_i - \hat{y}_i)^2} \quad (2)$$

$$MAE = \frac{1}{n}\sum_{i=1}^{n}|y_i - \hat{y}_i| \quad (3)$$

Where: $n$ is the number of samples, $y_i$ is the actual age and $\hat{y}$ is the predicted age.

*B. Results of Gender Classification*

Gender identification is a classification task where the primary measure used to assess the model is accuracy. Precision, recall, and F1-score, as well as the ROC curve, are also reported for a more comprehensive assessment. The confusion matrix makes it easier to calculate different metrics by providing in-depth information about the model's performance. The confusion matrix with and without normalization is shown in Figure 6, and the model's performance is highlighted by the corresponding statistics shown in Table 2. The ROC curve is shown in Figure 7, where an area under the curve of 0.95 denotes the efficacy of the model.

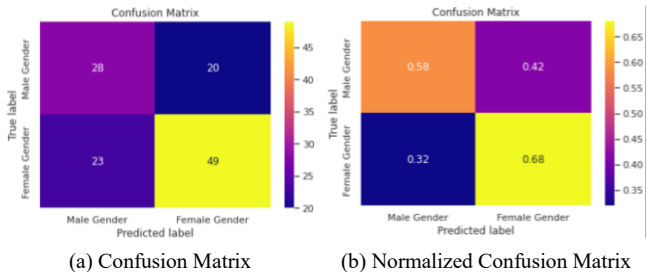

(a) Confusion Matrix     (b) Normalized Confusion Matrix

FIGURE 6: CONFUSION MATRIX FOR GENDER CLASSIFICATION

In Table 2, "Class" represents the gender category, while "Support" indicates the number of sample images used to compute the performance metric. The macro-averaged and weighted average performance for each metric is reported by averaging or weighted averaging across the support samples in each class.

TABLE 2 CLASSIFICATION PERFORMANCE OF GENDER CLASSIFICATION

| Class | Precision | Recall | F1-Score | Support |
|---|---|---|---|---|
| 0 | 0.55 | 0.58 | 0.57 | 48 |
| 1 | 0.71 | 0.68 | 0.7 | 72 |
| Accuracy | - | - | 0.64 | 120 |
| Macro-Average | 0.63 | 0.63 | 0.63 | 120 |
| Weighted-Average | 0.65 | 0.64 | 0.64 | 120 |

*C. Results of Age Estimation*

As a regression problem, age estimation requires the use of evaluation metrics such as RMSE and MAE to assess the model's performance. The model evaluation in this study is carried out using a dataset with 30% holdout. The evaluation results for the MAE and the RMSE are reported in Table 3.

TABLE 3: PERFORMANCE EVALUATION FOR AGE ESTIMATION

| Metric | Score |
|---|---|
| MSE | 52.529 |
| RMSE | 7.2477 |
| MAE | 5.7732 |

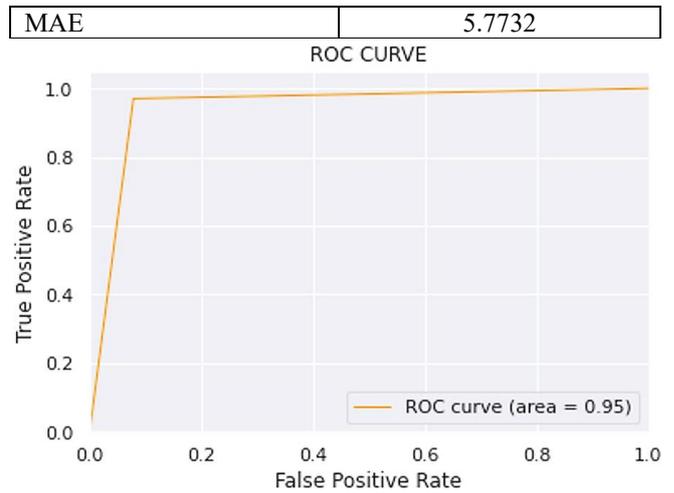

FIGURE 7: ROC CURVE AND AUC FOR GENDER CLASSIFICATION

*D. Discussion*

The results of the model evaluation demonstrate that the model performs well in both gender prediction and age estimation tasks. The mean absolute error for age estimation indicates that the model has an error of 5.77 years. Even though this error margin is small, further improvements are required to improve the precision of age estimation. By utilizing complementary information to increase age estimation precision, adding more modalities in addition to face images may greatly improve prediction accuracy.

Furthermore, the model's exceptional performance in predicting gender from face images is demonstrated by its 95% accuracy in gender identification. Even with the low error rate of 5%, there is still potential for improvement in terms of gender prediction precision. Notably, younger people have more subtle features that differentiate their faces from those of men and women, which may result in higher gender prediction error rates for this group. This emphasizes how crucial it is to keep improving and optimizing the model in order to attain even higher levels of gender identification accuracy, especially when dealing with younger subjects. It may be possible to further improve the model's predictive performance in age estimation and gender classification tasks by incorporating knowledge from the body of literature on feature extraction and model optimization methods.

V. CONCLUSION

This study concludes by presenting a deep learning-based method for classifying gender and estimating age from facial images. With a mean absolute error of 5.77 years for age estimation and a 95% accuracy rate for gender identification, the proposed approach shows promise in terms of accurate gender and age prediction. Even though these results show strong predictive abilities, there is still room for improvement, especially in terms of minimizing the age estimation error rate and improving the gender prediction accuracy, especially for younger people whose distinguishing characteristics might not be as prominent. In order to further improve prediction performance, future studies may investigate the integration of other modalities in addition to facial images. In general, this research advances methods for estimating age and gender, with potential uses in targeted advertising and other fields. Prospects for novel insights and uses in computer vision and machine learning are promising when research in this field is pursued further.

## REFERENCES

[1] Zebrowitz, L., *Reading faces: Window to the soul?* 2018: Routledge.
[2] Abayomi-Alli, A., et al. *Facial image quality assessment using an ensemble of pre-trained deep learning models (EFQnet)*. in *2020 20th international conference on computational science and its applications (ICCSA)*. 2020. IEEE.
[3] Gallagher, A.C. and T. Chen. *Estimating age, gender, and identity using first name priors*. in *2008 IEEE Conference on Computer Vision and Pattern Recognition*. 2008. IEEE.
[4] Fontanini, T., et al. *Automatic generation of semantic parts for face image synthesis*. in *International Conference on Image Analysis and Processing*. 2023. Springer.
[5] Ning, X., et al., *Multi-view frontal face image generation: a survey.* Concurrency and Computation: Practice and Experience, 2023. **35**(18): p. e6147.
[6] Hossain, S., et al., *Fine-grained image analysis for facial expression recognition using deep convolutional neural networks with bilinear pooling.* Applied Soft Computing, 2023. **134**: p. 109997.
[7] Siddiqi, M.H., et al., *Face image analysis using machine learning: a survey on recent trends and applications.* Electronics, 2022. **11**(8): p. 1210.
[8] George, P.A. and G.J. Hole, *Factors influencing the accuracy of age estimates of unfamiliar faces.* Perception, 1995. **24**(9): p. 1059-1073.
[9] Zimbler, M.S., M.S. Kokoska, and J.R. Thomas, *Anatomy and pathophysiology of facial aging.* Facial plastic surgery clinics of North America, 2001. **9**(2): p. 179-187.
[10] Fu, Y., G. Guo, and T.S. Huang, *Age synthesis and estimation via faces: A survey.* IEEE transactions on pattern analysis and machine intelligence, 2010. **32**(11): p. 1955-1976.
[11] Albert, A.M., K. Ricanek Jr, and E. Patterson, *A review of the literature on the aging adult skull and face: Implications for forensic science research and applications.* Forensic science international, 2007. **172**(1): p. 1-9.
[12] Aslam, M.A., N. Ahmed, and G. Saleem, *Vrl-iqa: visual representation learning for image quality assessment.* IEEE Access, 2023.
[13] Mark, L.S., et al., *Wrinkling and head shape as coordinated sources of age-level information.* Perception & Psychophysics, 1980. **27**(2): p. 117-124.
[14] Ahmed, N., et al., *Deep ensembling for perceptual image quality assessment.* Soft Computing, 2022. **26**(16): p. 7601-7622.
[15] Ahmed, N. and H.M.S. Asif, *Perceptual quality assessment of digital images using deep features.* Computing and Informatics, 2020. **39**(3): p. 385-409.
[16] Ahmed, N., H.M.S. Asif, and H. Khalid, *PIQI: perceptual image quality index based on ensemble of Gaussian process regression.* Multimedia Tools and Applications, 2021. **80**(10): p. 15677-15700.
[17] Park, U., Y. Tong, and A.K. Jain, *Age-invariant face recognition.* IEEE transactions on pattern analysis and machine intelligence, 2010. **32**(5): p. 947-954.
[18] Ahmad, N., et al., *Efficient jpeg encoding using bernoulli shift map for secure communication.* Wireless Personal Communications, 2022. **125**(4): p. 3405-3424.
[19] Ahmad, N., et al., *Leaf image-based plant disease identification using color and texture features.* Wireless Personal Communications, 2021. **121**(2): p. 1139-1168.
[20] Akhtar, M., *Automated analysis of visual leaf shape features for plant classification.* Computers and Electronics in Agriculture, 2019. **157**: p. 270-280.
[21] Maqsood, R., et al., *Anomaly recognition from surveillance videos using 3D convolution neural network.* Multimedia Tools and Applications, 2021.
[22] Saleem, G., et al., *Edge-Enhanced TempoFuseNet: A Two-Stream Framework for Intelligent Multiclass Video Anomaly Recognition in 5G and IoT Environments.* Future Internet, 2024. **16**(3): p. 83.
[23] Saleem, G., U.I. Bajwa, and R.H. Raza, *Toward human activity recognition: a survey.* Neural Computing and Applications, 2023. **35**(5): p. 4145-4182.
[24] Zaman, M.I., et al., *A robust deep networks based multi-object multi-camera tracking system for city scale traffic.* Multimedia Tools and Applications, 2023: p. 1-19.
[25] Saleem, G., U.I. Bajwa, and R.H. Raza, *Surveilia: Anomaly Identification Using Temporally Localized Surveillance Videos.* Available at SSRN 4308311.
[26] Fu, Y. and T.S. Huang, *Human age estimation with regression on discriminative aging manifold.* IEEE Transactions on Multimedia, 2008. **10**(4): p. 578-584.
[27] KHALIFA, T. and G. ŞENGÜL, *Gender prediction from facial images using local binary patterns and histograms of oriented gradients transformations.* Niğde Ömer Halisdemir Üniversitesi Mühendislik Bilimleri Dergisi, 2018. **7**(1): p. 14-22.
[28] Ahmed, N. and H.M.S. Asif. *Ensembling convolutional neural networks for perceptual image quality assessment*. in *2019 13th International Conference on Mathematics, Actuarial Science, Computer Science and Statistics (MACS)*. 2019. IEEE.
[29] Panner Selvam, I.R. and M. Karuppiah, *Gender recognition based on face image using reinforced local binary patterns.* IET Computer Vision, 2017. **11**(6): p. 415-425.
[30] Liu, X., et al., *Face image age estimation based on data augmentation and lightweight convolutional neural network.* Symmetry, 2020. **12**(1): p. 146.
[31] Li, P., et al., *Deep label refinement for age estimation.* Pattern Recognition, 2020. **100**: p. 107178.
[32] Guehairia, O., et al., *Feature fusion via Deep Random Forest for facial age estimation.* Neural Networks, 2020. **130**: p. 238-252.
[33] Aminian, A. and G. Noubir. *Deep Cross-Modal Age Estimation*. in *Advances in Computer Vision: Proceedings of the 2019 Computer Vision Conference (CVC), Volume 1 1*. 2020. Springer.
[34] Agbo-Ajala, O. and S. Viriri, *Deeply learned classifiers for age and gender predictions of unfiltered faces.* The Scientific World Journal, 2020. **2020**.
[35] Lim, S., *Estimation of gender and age using CNN-based face recognition algorithm.* International journal of advanced smart convergence, 2020. **9**(2): p. 203-211.
[36] Yaman, D., F. Irem Eyiokur, and H. Kemal Ekenel. *Multimodal age and gender classification using ear and profile face images*. in *Proceedings of the IEEE/CVF Conference on Computer Vision and Pattern Recognition Workshops*. 2019.
[37] Wirth, R. and J. Hipp. *CRISP-DM: Towards a standard process model for data mining*. in *Proceedings of the 4th international conference on the practical applications of knowledge discovery and data mining*. 2000. Manchester.